\def\year{2022}\relax
\documentclass[letterpaper]{article} 
\usepackage{aaai22} 
\usepackage{times} 
\usepackage{helvet} 
\usepackage{courier} 
\usepackage[hyphens]{url} 
\usepackage{graphicx} 
\usepackage{natbib} 
\usepackage{caption} 
\DeclareCaptionStyle{ruled}{labelfont=normalfont,labelsep=colon,strut=off} 
\usepackage{adjustbox}
\usepackage{algorithm}
\usepackage{algorithmic}

\usepackage{todonotes}
\usepackage{hyperref}
\usepackage{booktabs}
\usepackage{newfloat}
\usepackage{multirow}
\usepackage{listings}
\usepackage{subcaption}

\floatstyle{ruled}
\newfloat{listing}{tb}{lst}{}
\floatname{listing}{Listing}
\title{Evaluation of Fake News Detection with Knowledge-Enhanced Language Models}
\author{
Chenxi Whitehouse, Tillman Weyde, Pranava Madhyastha, Nikos Komninos
}
\affiliations{
City, University of London

\{chenxi.whitehouse, t.e.weyde, pranava.madhyastha, nikos.komninos.1\}@city.ac.uk

}
\begin{document}

\maketitle

\begin{abstract}

Recent advances in fake news detection have exploited the success of large-scale pre-trained language models (PLMs).
The predominant state-of-the-art approaches are based on fine-tuning PLMs on labelled fake news datasets. 
However, large-scale PLMs are generally not trained on structured factual data and hence may not possess priors that are grounded in factually accurate knowledge. 
The use of existing knowledge bases (KBs) with rich human-curated factual information has thus the potential to make fake news detection more effective and robust.  
In this paper, we investigate the impact of knowledge integration into PLMs for fake news detection. 
We study several state-of-the-art approaches for knowledge integration, mostly using Wikidata as KB, on two popular fake news datasets - \texttt{LIAR}, a politics-based dataset, and \texttt{COVID-19}, a dataset of messages posted on social media relating to the COVID-19 pandemic. 
Our experiments show that knowledge-enhanced models can significantly improve fake news detection on \texttt{LIAR} where the KB is relevant and up-to-date.
The mixed results on \texttt{COVID-19} highlight the reliance on stylistic features and the importance of domain-specific and current KBs. 
The code is available at \url{https://github.com/chenxwh/fake-news-detection}.

\end{abstract}

\section{Introduction}
The world is witnessing a growing epidemic of fake news, which includes misinformation, disinformation, rumours, hoaxes, and other forms of rapid spread and factually inaccurate information~\cite{Sharma2019CombatingFN}. 
Fake news has been observed to severely impact political processes because of the wide reach of social media~\cite{2016-Election}. 
Misinformation related to medical issues, such as the COVID-19 pandemic, can cost lives~\cite{OConnor2020GoingVD}.
Automated methods for fake news detection and mitigation are a critical yet technically challenging problem~\cite{thorne-vlachos-2018-automated}.

In this paper, we focus on content-based fake news detection: methods that assess the truthfulness of news items based only on the text without using metadata. 
State-of-the-art models for this task are driven by advances in 
large-scale pre-trained language models (PLMs)~\citep[e.g.][]{Liu2019ATM, Kaliyar2021FakeBERT}, which are trained on vast amounts of raw web-based text using self-supervised methods~\cite{rogers-etal-2020-primer}.  
A major limitation of these models is the lack of explicit grounding to real-world entities and relations, which makes it difficult to recover factual knowledge~\cite{Bender2021OnTD}. 
On the other hand, knowledge bases (KBs) provide a rich source of structured and human-curated factual knowledge, often complementary to what is found in raw text. 
This has recently led to the development of KB-augmented language models. 
Fake news detection can particularly benefit from the integration of KBs, making such models less dependent and reliant on surface-level linguistic features. 

In this study, we empirically analyse the impact of recent state-of-the-art knowledge integration methods, which enhance PLMs with KBs, for content-based fake news detection tasks. 
We evaluate ERNIE~\cite{zhang-etal-2019-ernie},  KnowBert~\cite{peters-etal-2019-knowledge}, KEPLER~\cite{kepler} and K-ADAPTER~\cite{wang-etal-2021-k} on two distinct publicly available datasets:  \texttt{LIAR}~\cite{wang-2017-liar}, a politically oriented dataset, and \texttt{COVID-19}~\cite{Patwa2021FightingAI}, a dataset related to the recent pandemic. 
We find that integrating knowledge can improve fake news detection accuracy, given that the knowledge bases are relevant and up-to-date.
Our experiments are not designed to find new state-of-the-art models for these datasets, but to investigate the effect of knowledge base integration into PLMs. 

Our contributions are as follows: we evaluate multiple KB integration methods for fake news detection, we investigate model and data  aspects that can prevent KB integration from being effective or from being effectively measured, and we discuss the potential for real-world applications.

In the following sections, we present a brief overview of four state-of-the-art methods that integrate KBs with PLMs studied in this paper.
We then introduce and compare the datasets, the experiments with different knowledge-enhanced models, and the effectiveness of entity linking.
We discuss our findings with respect to the necessary conditions for KB integration to be effective and how to assess its effect in application scenarios.  
Finally, we discuss the challenges in fake news detection and promising future directions.

\section{Method}

In this section, we introduce the models with KB integration and describe the datasets and our experimental setup. 

\subsection{Knowledge Integration for PLM}
Standard deep learning models obtain information from predicting and classifying text as they are trained, but have no prior knowledge of, or interaction with, world knowledge. 
Although PLMs can effectively characterise linguistic patterns from text to generate high-quality context-aware representations, they are limited in their grasp of knowledge, concepts, and relations, which are essential for some Natural Language Processing (NLP) tasks, including assessing the truthfulness of news items.

On the other hand, KBs like Wikidata (\url{https://www.wikidata.org}) and WordNet~\cite{Miller1995WordNetAL} contain rich curated information about the world. 
Thus, they could greatly complement PLMs if effective integration methods were available. 
Several efforts have been made to integrate KBs into PLMs. 
In this paper, we study the following models:


\begin{figure*}
\centering
\adjustbox{max width=0.98\textwidth}{%
\centering
\begin{subfigure}[b]{0.265\textwidth}
\centering
\includegraphics[width=\textwidth]{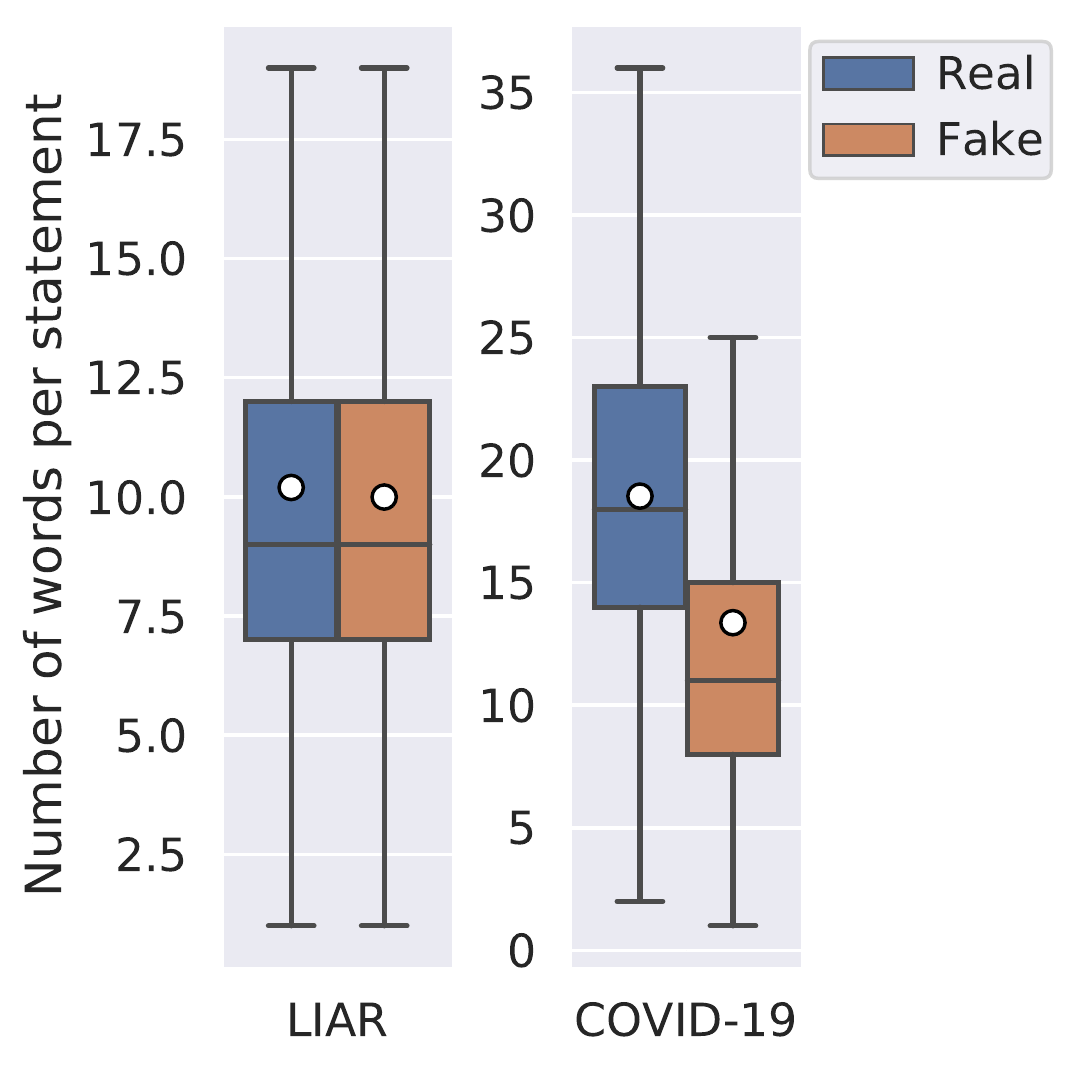}
\caption{Word count per statement}
\label{fig:len}
\end{subfigure}
\hfill
\begin{subfigure}[b]{0.345\textwidth}
\centering
\includegraphics[width=\textwidth]{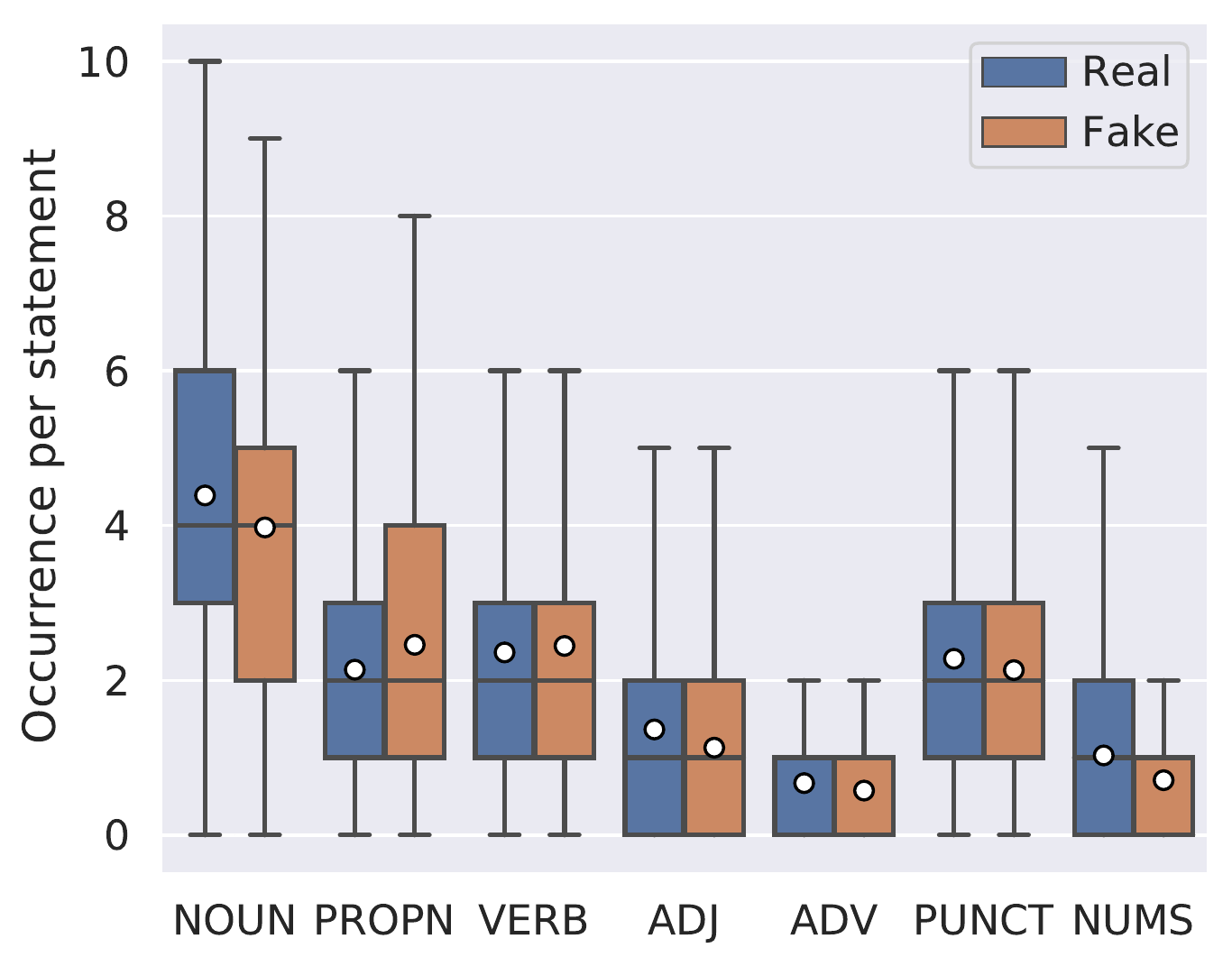}
\caption{POS, punctuation, numbers in \texttt{LIAR}}
\label{fig:liarPos}
\end{subfigure}
\hfill
\begin{subfigure}[b]{0.38\textwidth}
\centering
\includegraphics[width=\textwidth]{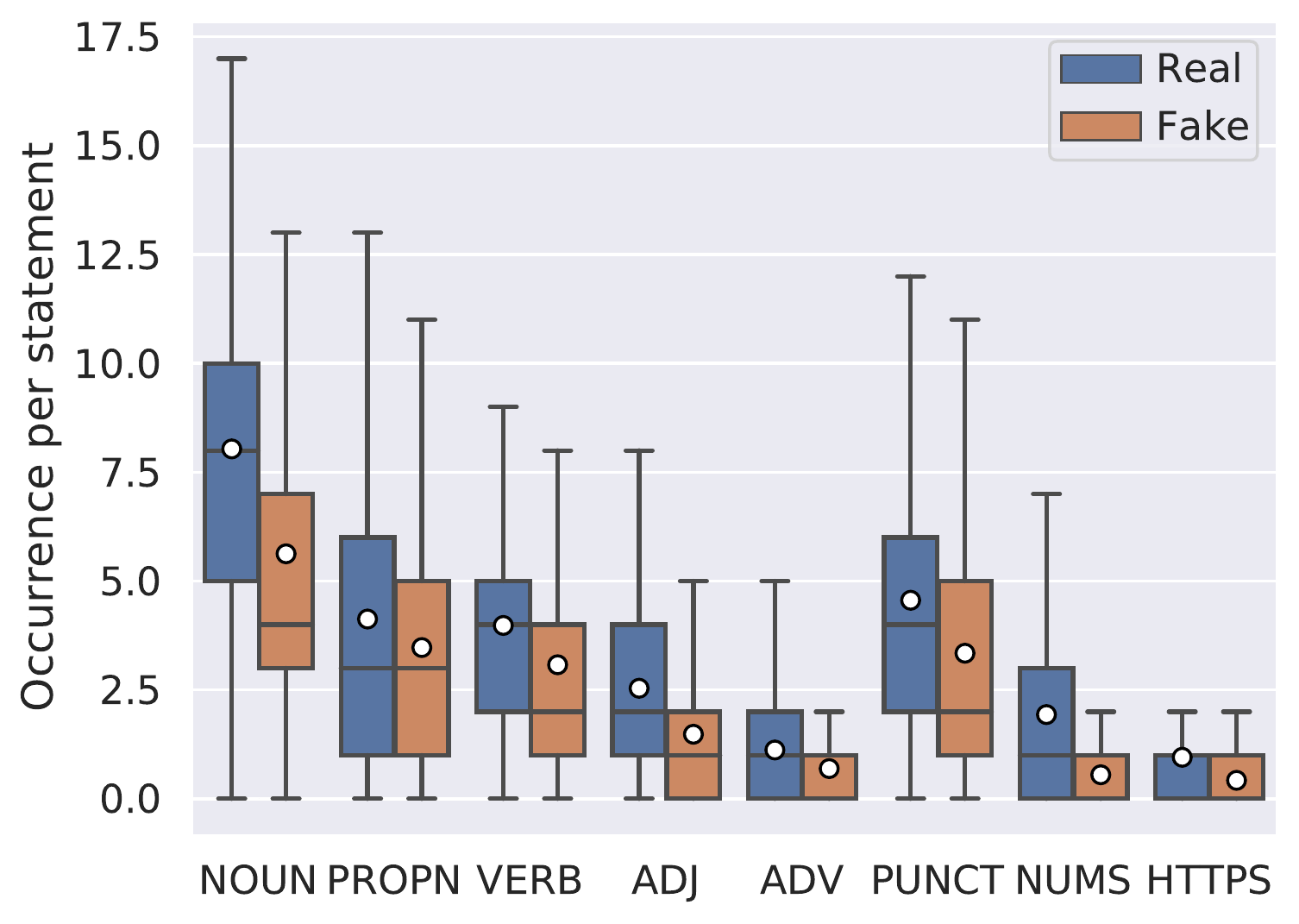}
\caption{POS, punctuation, numbers, https in \texttt{COVID-19}}
\label{fig:covidPos}
\end{subfigure}
}
\caption{Number of words, POS tags, punctuation and numbers per statement in real and fake news in \texttt{LIAR} and \texttt{COVID-19}, and
number of https-links per statement in \texttt{COVID-19}.
The mean values are shown as white-filled circles in the plot.}
\label{fig:box_plot}
\end{figure*}


\paragraph{ERNIE} injects knowledge into BERT \cite{devlin-etal-2019-bert} by pre-training a language model on both large corpora and KBs. 
It uses TAGME \cite{Ferragina2010TAGMEOA} to link entities to Wikidata.
TAMGE identifies entity mentions in the input text and links them to associated entity embeddings, which are then fused into the corresponding positions of the text.
The knowledge-based learning objective is to predict the correct token-entity alignment. 
ERNIE has enhanced performance over BERT in entity typing and relation classification~\cite{zhang-etal-2019-ernie}. 

\paragraph{KnowBert} incorporates KBs into BERT using a knowledge attention and re-contextualisation mechanism.
It identifies mention spans in the input text and incorporates an integrated entity linker to retrieve entity embeddings from a KB. 
The entity linker is responsible for entity disambiguation, which considers 30 entity candidates and uses their weighted average embedding. 
Knowledge-enhanced entity-span representations are then re-contextualised with a word-to-entity attention technique. 
KnowBert has shown improvement over BERT in relationship extraction, entity typing and word sense disambiguation \cite{peters-etal-2019-knowledge}.

\paragraph{KEPLER} integrates factual knowledge into PLMs by adding a knowledge embedding objective with the supervision from a KB and optimising it jointly with language modelling objectives.
KEPLER is trained to encode the entities from their contextual descriptions, which enhances the ability of PLMs to extract knowledge from text.
By keeping the original structures of PLMs, KEPLER can be used in general downstream NLP tasks without additional inference overhead.
It is shown that KEPLER improves performance over RoBERTa \cite{Liu2019RoBERTaAR} in relationship extraction, entity typing and link prediction~\cite{kepler}. 

\paragraph{K-ADAPTER} retains the PLMs unchanged, 
but adds learnable adapter features that are trained in a multi-task setting on relation prediction and dependency-tree prediction. 
Two kinds of knowledge adapters have been developed by \citet{wang-etal-2021-k}: factual knowledge obtained from automatically aligned text triples on Wikipedia and Wikidata, and linguistic knowledge obtained via dependency parsing. 
Both have been found to improve relation classification, entity typing and question answering~\cite{wang-etal-2021-k}.

\subsection{Datasets}
In our experiments, we use \texttt{LIAR} and \texttt{COVID-19} to study fake news detection.
They both consist of short statements, but with different content, time of collection, and linguistic and stylistic features.

\subsubsection{\texttt{LIAR}} was collected in 2017 from Politifact (\url{https://www.politifact.com}).
It includes 12.8k human-labelled short statements about US politics from various contexts, i.e. news releases, TV interviews, campaign speeches, etc. 
Each statement has been rated for truthfulness by a Politifact editor using a six-grade scale: ``pants-fire", ``false", ``barely-true", ``half-true", ``mostly true", and ``true".
\texttt{LIAR} also provides metadata (e.g. speaker, context),
which we do not use in our experiments.
While~\citet{wang-2017-liar} has been widely cited, 
we only found three other results for our specific task (no metadata, six classes):~\cite{alhindi-etal-2018-evidence, Liu2019ATM, Chernyavskiy2020RecursiveNT}, the latter has the current best accuracy of 34.5\%.
\subsubsection{\texttt{COVID-19}} was collected in 2020 after the COVID-19 outbreak. 
It consists of 10.5k posts related to the pandemic which are obtained from different social media sites including Twitter, Facebook, and Instagram. 
The fake posts were collected from various fact-checking websites, i.e. Politifact and NewsChecker (\url{https://newschecker.in}), and the real posts were from Twitter using verified Twitter handles. 
Each post has a label, ``real" or ``fake".
It was used as a shared task in the CONSTRAINT 2021 workshop~\cite{chakraborty2021combating} with the best-reported accuracy of~98.69\%.

\subsection{Experimental Setup}
We use an empirical approach to study the effect of knowledge integration on fake news detection, to understand how knowledge is used by the model, and to evaluate the quality of the entity linker to the KB.

ERNIE and KnowBert are built on BERT-base, whereas KEPLER and K-ADAPTER are enhanced from RoBERTa-base and RoBERTa-large, respectively.
We follow the concept of an ablation study to investigate the influence of external knowledge by comparing the performance of each knowledge-enhanced PLM with the corresponding baseline model.
We note that ERNIE and KnowBert incorporate entity embeddings that are linked to the input.
Therefore we visualise the entities linked that contribute to the fake news detection task in ERNIE, and design experiments to investigate the impact of entity disambiguation of KnowBert.

We evaluate the performance of the models on fake news detection by fine-tuning the knowledge-enhanced PLMs on the training set with the same hyperparameter settings.
The input text is fed first to the PLM, and followed by a dropout ($p=0.1$) and a linear layer. The output is then passed to a softmax layer for classification.
We use AdamW optimiser
\cite{Loshchilov_Hutter_2019_Decoupled} (learning rate of $5\times10^{-6}$) and cross entropy as the loss function.
The maximum input length is set to 128, and the batch size is 4. 
We train for 10 epochs and usually observe convergence after five.
We perform five runs for each experiment and report the average accuracy with the standard deviation.  
Both \texttt{LIAR} and \texttt{COVID-19} are already divided into train, validation, and test splits, which we use in our experiments as provided.

\subsubsection{Linguistic Feature Analysis}
We also perform linguistic feature analysis following the work in 
\citet{Horne_Adali_2017} to investigate the stylistic differences between real and fake news in the datasets.
We use spaCy (\url{https://spacy.io}) to parse the statements and get the Part-of-Speech (POS) tags. 
For \texttt{LIAR}, we group ``pants-fire", ``false", and ``barely-true" as fake and ``half-true", ``mostly true", and ``true" as real. 
We compare the distribution of different words, POS tags (NOUN, PROPN, VERB, ADJ, ADV), punctuation, and number-like words in each statement in Figure \ref{fig:box_plot}.

The length of posts is quite different between the two classes in \texttt{COVID-19}, with an average of 32 and 22 for real and fake statements, respectively, as shown in  
\autoref{fig:len}. 
\texttt{LIAR}, on the other hand, has a similar statement length, with 18 words per statement for real and 17 for fake.

In general, \texttt{COVID-19} has distinct linguistic features between classes whereas \texttt{LIAR} shows more similar features. 
In particular, \texttt{COVID-19} contains links, mostly https links, which are listed as a separate category in \autoref{fig:covidPos}, showing a very skewed distribution.

\section{Experiments and Results}

For our experiments, we use ERNIE, three pre-trained KnowBert models with different KBs (Wiki, WordNet, W+W), KEPLER, and K-ADAPTER with three adapters (F, L, F-L) in the published implementation, fine-tune the models to our task and compare the result with the baseline models - BERT-base, RoBERTa-base, and RoBERTa-large.

\paragraph{Detection Accuracy}

\begin{table}[tb]
\centering
\adjustbox{max width=0.9\linewidth}{%
\begin{tabular}{lclc}
\toprule
\sc {Model} & \sc {Base} & \multicolumn{1}{c}{ \sc {LIAR}} &\sc { COVID-19} \\
\midrule
 \textbf{B}ERT-\textbf{B}ase (BB) & - & 26.36 \textsubscript{$\pm$0.58} & 97.51 \textsubscript{$\pm$0.19} \\
\textbf{R}oBERTa-\textbf{B}ase (RB) & -  & 26.71 \textsubscript{$\pm$0.93} & 97.61 \textsubscript{$\pm$0.26}\\
\textbf{R}oBERTa-\textbf{L}arge (RL) & -  & \textbf{27.36} \textsubscript{$\pm$0.79} & \textbf{97.92} \textsubscript{$\pm$0.17}\\ 
 \midrule
ERNIE & BB &27.53 \textsubscript{$\pm$0.13} & 97.30 \textsubscript{$\pm$0.18} \\
 KnowBert-Wiki & BB & 27.64 \textsubscript{$\pm$0.09} & 97.37 \textsubscript{$\pm$0.09} \\ 
 KEPLER & RB & 26.77 \textsubscript{$\pm$1.15} & 97.58 \textsubscript{$\pm$0.15} \\
K-ADAPTER-F & RL &\textbf{28.63} \textsubscript{$\pm$0.90}$^*$ & \textbf{97.92} \textsubscript{$\pm$0.10}\\
\midrule
  KnowBert-WordNet & BB  & 26.95 \textsubscript{$\pm$0.45} & 97.00 \textsubscript {$\pm$0.06}\\
 KnowBert-W+W & BB  & \textbf{28.95} \textsubscript{$\pm$0.64}$^*$ & 97.56 \textsubscript{$\pm$0.15}\\ 
  K-ADAPTER-L & RL & 28.46 \textsubscript{$\pm$0.87}$^*$  & 98.07 \textsubscript{$\pm$0.09} \\
  K-ADAPTER-F-L & RL & 27.45 \textsubscript{$\pm$0.78} & \textbf{98.11} \textsubscript{$\pm$0.14} \\  
  \bottomrule
\end{tabular}
}
 \caption{Detection accuracy results (average of five runs).
 The first section corresponds to the baseline models. 
 Models in the second section use Wikidata KB. 
 The third section shows models using other KBs and features.  
 The best values within each section per dataset are marked in bold.
 The subscript numbers with $\pm$ show the standard deviation.
 Results with $^*$ indicate statistically significant improvements over the baseline, both for the mean (t-test, one-sided, $p<.05$) and median (Wilcoxon signed rank test, one-sided, $p<.05$). 
 }
 \label{tab:acc_table}
\end{table}

The detection accuracy of the knowledge-enhanced PLMs and the corresponding baselines is shown in \autoref{tab:acc_table}. 
On \texttt{LIAR}, all knowledge-enhanced methods improve over the baseline with KnowBert-W+W reaching the best overall result (improvement of $+2.59$ over BERT-base), whereas, on \texttt{COVID-19}, only 
three of eight models show improvement, and only by a small margin.

The computational cost varies per approach. 
KEPLER retains the baseline PLM architecture, thus there is no overhead compared to RoBERTa-base.
K-ADAPTER also freezes the RoBERTa-large layers, but there is an overhead of 9-23\% from the adapters, while the overhead for KnowBert is 40-87\% and 111-131\% for ERNIE.

\begin{figure}
\centering
\begin{subfigure}[b]{0.46\textwidth}
\centering
\includegraphics[width=\textwidth]{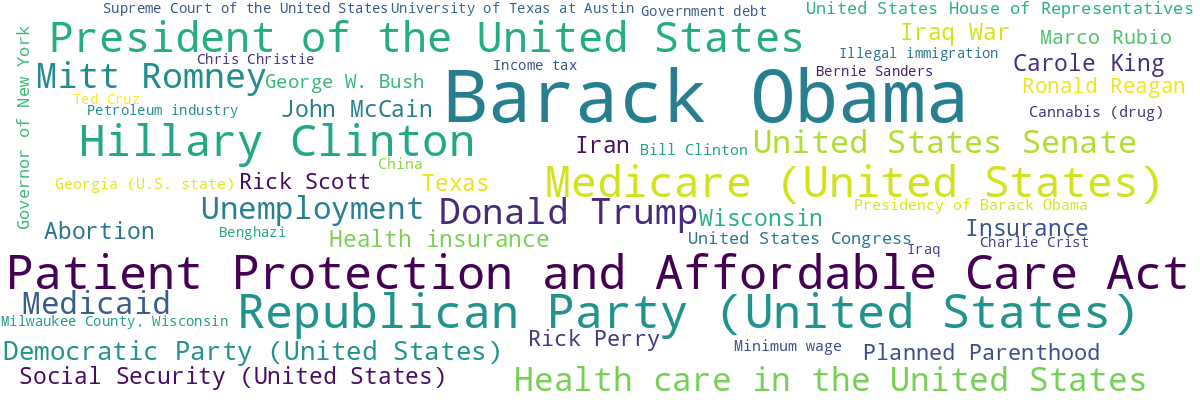}
\label{fig:e1}
\end{subfigure}
\hfill
\begin{subfigure}[b]{0.46\textwidth}

\includegraphics[width=\textwidth]{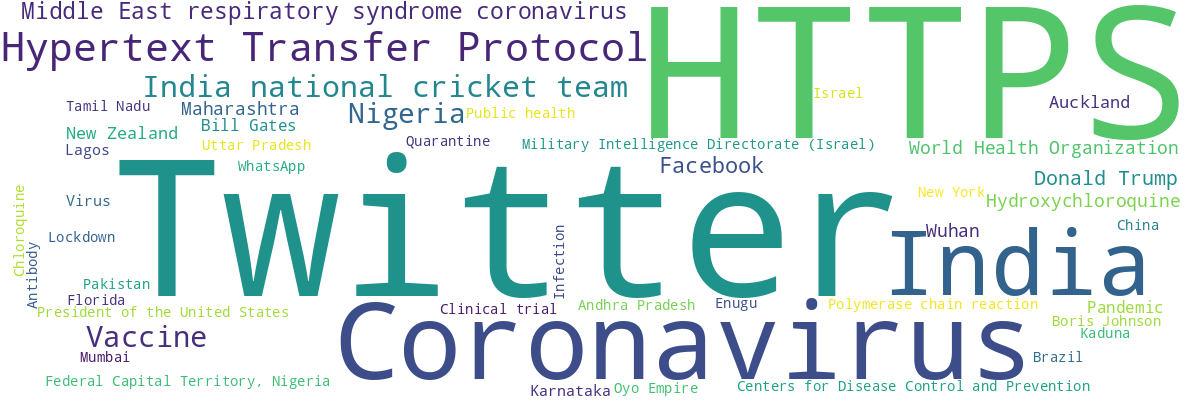}
\label{fig:e2}
\end{subfigure}
\caption{Word clouds for the 50 most frequent entities linked by ERNIE in  \texttt{LIAR} (top) and  \texttt{COVID-19} (bottom).}
\label{fig:word_cloud_Ernie}
\end{figure}

\paragraph{KB Linking}
ERNIE and KnowBert create links between the text and KB entities at runtime and the quality of this linking influences the output. 
ERNIE uses TAGME and selects only one entity candidate per text span. 
In \autoref{fig:word_cloud_Ernie} we show the 50  most frequently selected KB entities for each dataset.
We can see that in \texttt{COVID-19}, the most frequent entities are not content-related (``https", ``twitter") while ``COVID-19", the most frequent relevant term in the dataset, is missing in the linked entities. 
For \texttt{LIAR}, on the other hand, the linked entities seem relevant. 
Since \texttt{LIAR} was collected three years earlier, it is apparently a better match for the entity linker and the KB used.
Another potential influence on the effectiveness of KB integration is the number of linked entities.  
In contrast to ERNIE, KnowBert selects the 30 most probable entities per text span. 
In a sensitivity study, we restrict KnowBert-W+W to only one entity, which reduces the accuracy on \texttt{LIAR} from 28.95\% to 27.31\%, below the accuracy of ERNIE (27.53\%). 


\section{Discussion}

The reliable improvement of detection accuracy on \texttt{LIAR} by integrating PLMs with Wikidata shows the potential of knowledge integration exceeding the results obtained by integrating multiple types of metadata by \citet{wang-2017-liar}. 
On the other hand, the improvements are good but not dramatic for \texttt{LIAR} and not consistent for \texttt{COVID-19}. 
We can identify two aspects contributing to the result which are relevant to the effective use of knowledge-enhanced models: 

1) Currentness and relevance of the KB: as \texttt{COVID-19} was collected after most of the PLMs were trained, some terms such as ``COVID-19" are not in the KB; 

2) Quality of the dataset: 
the \texttt{COVID-19} dataset contains confounders that provide strong cues, overshadowing the impact of the knowledge base. 
The most important one is the occurrence of https links, which appear in 95.3\% of the real posts but only 42.3\% of the fake posts. 

There is also potential to achieve more explainability and interpretability with direct KB integration at runtime. 
Take this statement from \texttt{COVID-19}: \textit{``DNA Vaccine: injecting genetic material into the host so that host cells create proteins that are similar to those in the virus against which the host then creates antibodies"} as an example,
KnowBert-W+W correctly classifies it as ``real", whereas BERT-base fails.
We observe most mention spans in the statement, i.e.  \textit{``DNA"}, \textit{``injecting"}, \textit{``genetic"}, \textit{``genetic material"}, \textit{``host"}, \textit{``cells"}, etc. are correctly linked to entities \textit{``DNA"}, \textit{``Injection\_(medicine)"}, \textit{``Genetics"}, \textit{``Genome"}, \textit{``Host\_(biology)"}, \textit{``Cell\_(biology)"}, respectively, 
therefore it seems that the entity links may have contributed to KnowBert-W-W for this classification. 
However, the level of explainability is still limited.

\paragraph{Application Aspects}
Automatic fake news detection in practice adds two dynamic application aspects, which are difficult to test with static datasets as our experiment on \texttt{COVID-19} has shown:

(1) Dynamic adaptation: it is necessary to update the system to the changing characteristics of real and fake news~\cite{silva2021concept}.
Knowledge-enhanced models that use KBs at runtime offer an opportunity to update the KB independent of the model. %
This has the advantage that fake news can be recognised as contradicting the KB before there are any fake news examples. 

(2) Adversarial robustness: fake news authors are very likely to take evasive action. 
Adapting the text style is relatively easy and could be automated, which makes the detection with stylistic features difficult \citep[see][]{NEURIPS2019_3e9f0fc9, schuster-etal-2020-limitations}. 

Deployment of fake news detection in social media will also need human verification, e.g. when a user challenges actions taken against them. 
Here, KB integration can offer the advantage of insight into knowledge that has been used in the detection for better explainability.

\section{Related Work}

In recent years large-scale PLMs i.e. BERT and RoBERTa have dominated NLP tasks, including some content-based fake news detection~\cite{Kaliyar2021FakeBERT}. 
Most fake news detection approaches either combine text with metadata \citep[e.g.][]{ding2020bert} or focus only on the source of the text \citep[e.g.][]{gruppi2022nela}.
For \texttt{LIAR}, \citet{alhindi-etal-2018-evidence} extend the data with evidence sentences in a new dataset called \texttt{LIAR-PLUS} to improve detection.
\citet{Chernyavskiy2020RecursiveNT} introduce a Deep Averaging Network to model the discursive structure of the text and use Siamese models on the extended text data. 
\citet{Liu2019ATM} predict labels at two levels of granularity.
For \texttt{COVID-19}, there are a number of results from the CONSTRAINTS 2021 workshop~\cite{chakraborty2021combating} which use a wide variety of traditional and neural NLP models. 
None of these approaches uses external knowledge, so they could all benefit from KB integration.

\section{Conclusion and Future Work}

In this paper, we study the effectiveness of enhancing PLMs with knowledge bases for fake news detection. 
We find that integrating knowledge with PLMs can be beneficial on a static dataset but it depends on suitable KBs and the quality of the data. 
On the modelling level, there are many routes for improvement. 
For practical application, more insight into what knowledge is used would be useful as well as dynamic adaptation of the models and the KBs. 
Integrating KBs with PLMs offers potentially more robust and timely fake news detection. However, a new evaluation approach, i.e. a testing scenario that models dynamic knowledge as well as adversarial and automatic fake news generators, is needed to assess the true potential of knowledge integration.

\bibliography{mylib}

\begin{thebibliography}{28}
\providecommand{\natexlab}[1]{#1}

\bibitem[{Alhindi, Petridis, and Muresan(2018)}]{alhindi-etal-2018-evidence}
Alhindi, T.; Petridis, S.; and Muresan, S. 2018.
\newblock Where is Your Evidence: Improving Fact-checking by Justification
  Modeling.
\newblock In \emph{Proceedings of the First Workshop on Fact Extraction and
  {VER}ification ({FEVER})}, 85--90. Brussels, Belgium: ACL.

\bibitem[{Allcott and Gentzkow(2017)}]{2016-Election}
Allcott, H.; and Gentzkow, M. 2017.
\newblock Social Media and Fake News in the 2016 Election.
\newblock \emph{Journal of Economic Perspectives}, 31(2): 211--36.

\bibitem[{Bender et~al.(2021)Bender, Gebru, McMillan-Major, and
  Shmitchell}]{Bender2021OnTD}
Bender, E.~M.; Gebru, T.; McMillan-Major, A.; and Shmitchell, S. 2021.
\newblock On the Dangers of Stochastic Parrots: Can Language Models Be Too Big?
\newblock In \emph{Proceedings of the 2021 ACM Conference on Fairness,
  Accountability, and Transparency}, 610--623.

\bibitem[{Chakraborty(2021)}]{chakraborty2021combating}
Chakraborty, T. 2021.
\newblock \emph{Combating Online Hostile Posts in Regional Languages during
  Emergency Situation: First International Workshop, CONSTRAINT 2021,
  Collocated with AAAI 2021, Virtual Event, February 8, 2021, Revised Selected
  Papers}.
\newblock Springer Nature.

\bibitem[{Chernyavskiy and Ilvovsky(2020)}]{Chernyavskiy2020RecursiveNT}
Chernyavskiy, A.; and Ilvovsky, D. 2020.
\newblock Recursive Neural Text Classification Using Discourse Tree Structure
  for Argumentation Mining and Sentiment Analysis Tasks.
\newblock In \emph{ISMIS}, 90--101. Springer.

\bibitem[{Devlin et~al.(2019)Devlin, Chang, Lee, and
  Toutanova}]{devlin-etal-2019-bert}
Devlin, J.; Chang, M.-W.; Lee, K.; and Toutanova, K. 2019.
\newblock BERT: Pre-training of Deep Bidirectional Transformers for Language
  Understanding.
\newblock In \emph{NAACL-HLT}, 4171--4186. Minneapolis, Minnesota: ACL.

\bibitem[{Ding, Hu, and Chang(2020)}]{ding2020bert}
Ding, J.; Hu, Y.; and Chang, H. 2020.
\newblock BERT-based Mental Model, a Better Fake News Detector.
\newblock In \emph{Proceedings of the 2020 6th international conference on
  computing and artificial intelligence}, 396--400.

\bibitem[{Ferragina and Scaiella(2010)}]{Ferragina2010TAGMEOA}
Ferragina, P.; and Scaiella, U. 2010.
\newblock TAGME: On-the-fly Annotation of Short Text Fragments (by Wikipedia
  Entities).
\newblock \emph{Proceedings of the 19th ACM international conference on
  Information and knowledge management}.

\bibitem[{Horne and Adali(2017)}]{Horne_Adali_2017}
Horne, B.; and Adali, S. 2017.
\newblock This Just In: Fake News Packs A Lot in Title, Uses Simpler,
  Repetitive Content in Text Body, More Similar to Satire Than Real News.
\newblock \emph{Proceedings of the International AAAI Conference on Web and
  Social Media}, 11(1): 759--766.

\bibitem[{Kaliyar, Goswami, and Narang(2021)}]{Kaliyar2021FakeBERT}
Kaliyar, R.~K.; Goswami, A.; and Narang, P. 2021.
\newblock FakeBERT: Fake News Detection in Social Media with a BERT-based Deep
  Learning Approach.
\newblock \emph{Multimedia tools and applications}, 80(8): 11765--11788.

\bibitem[{Liu et~al.(2019{\natexlab{a}})Liu, Wu, Yu, Li, Jiang, qing Huang, and
  Lu}]{Liu2019ATM}
Liu, C.; Wu, X.; Yu, M.; Li, G.; Jiang, J.; qing Huang, W.; and Lu, X.
  2019{\natexlab{a}}.
\newblock A Two-Stage Model Based on BERT for Short Fake News Detection.
\newblock In \emph{KSEM}.

\bibitem[{Liu et~al.(2019{\natexlab{b}})Liu, Ott, Goyal, Du, Joshi, Chen, Levy,
  Lewis, Zettlemoyer, and Stoyanov}]{Liu2019RoBERTaAR}
Liu, Y.; Ott, M.; Goyal, N.; Du, J.; Joshi, M.; Chen, D.; Levy, O.; Lewis, M.;
  Zettlemoyer, L.; and Stoyanov, V. 2019{\natexlab{b}}.
\newblock RoBERTa: A Robustly Optimized BERT Pretraining Approach.
\newblock \emph{CoRR}, abs/1907.11692.

\bibitem[{Loshchilov and Hutter(2019)}]{Loshchilov_Hutter_2019_Decoupled}
Loshchilov, I.; and Hutter, F. 2019.
\newblock Decoupled Weight Decay Regularization.
\newblock In \emph{7th International Conference on Learning Representations,
  {ICLR} 2019}.

\bibitem[{Miller(1995)}]{Miller1995WordNetAL}
Miller, G. 1995.
\newblock WordNet: A Lexical Database for English.
\newblock \emph{Commun. ACM}, 38: 39--41.

\bibitem[{Nørregaard, Horne, and Adali(2019)}]{gruppi2022nela}
Nørregaard, J.; Horne, B.~D.; and Adali, S. 2019.
\newblock NELA-GT-2018: A Large Multi-Labelled News Dataset for the Study of
  Misinformation in News Articles.
\newblock \emph{Proceedings of the International AAAI Conference on Web and
  Social Media}, 13(01): 630--638.

\bibitem[{O’Connor and Murphy(2020)}]{OConnor2020GoingVD}
O’Connor, C.; and Murphy, M. 2020.
\newblock Going Viral: Doctors Must Tackle Fake News in the Covid-19 Pandemic.
\newblock \emph{Bmj}, 369(10.1136).

\bibitem[{Patwa et~al.(2021)Patwa, Sharma, Pykl, Guptha, Kumari, Akhtar, Ekbal,
  Das, and Chakraborty}]{Patwa2021FightingAI}
Patwa, P.; Sharma, S.; Pykl, S.; Guptha, V.; Kumari, G.; Akhtar, M.~S.; Ekbal,
  A.; Das, A.; and Chakraborty, T. 2021.
\newblock Fighting an Infodemic: COVID-19 Fake News Dataset.
\newblock In \emph{CONSTRAINT@AAAI}.

\bibitem[{Peters et~al.(2019)Peters, Neumann, Logan, Schwartz, Joshi, Singh,
  and Smith}]{peters-etal-2019-knowledge}
Peters, M.~E.; Neumann, M.; Logan, R.; Schwartz, R.; Joshi, V.; Singh, S.; and
  Smith, N.~A. 2019.
\newblock Knowledge Enhanced Contextual Word Representations.
\newblock In \emph{EMNLP/IJCNLP}, 6086--6093. Hong Kong, China: ACL.

\bibitem[{Rogers, Kovaleva, and Rumshisky(2020)}]{rogers-etal-2020-primer}
Rogers, A.; Kovaleva, O.; and Rumshisky, A. 2020.
\newblock A Primer in {BERT}ology: What We Know about How {BERT} Works.
\newblock \emph{Transactions of the ACL}, 8: 842--866.

\bibitem[{Schuster et~al.(2020)Schuster, Schuster, Shah, and
  Barzilay}]{schuster-etal-2020-limitations}
Schuster, T.; Schuster, R.; Shah, D.~J.; and Barzilay, R. 2020.
\newblock The Limitations of Stylometry for Detecting Machine-Generated Fake
  News.
\newblock \emph{Computational Linguistics}, 46(2): 499--510.

\bibitem[{Sharma et~al.(2019)Sharma, Qian, Jiang, Ruchansky, Zhang, and
  Liu}]{Sharma2019CombatingFN}
Sharma, K.; Qian, F.; Jiang, H.; Ruchansky, N.; Zhang, M.; and Liu, Y. 2019.
\newblock Combating Fake News.
\newblock \emph{ACM Transactions on Intelligent Systems and Technology (TIST)},
  10: 1 -- 42.

\bibitem[{Silva and Almeida(2021)}]{silva2021concept}
Silva, R.~M.; and Almeida, T.~A. 2021.
\newblock How Concept Drift can Impair the Classification of Fake News.
\newblock In \emph{Anais do IX Symposium on Knowledge Discovery, Mining and
  Learning}, 121--128. SBC.

\bibitem[{Thorne and Vlachos(2018)}]{thorne-vlachos-2018-automated}
Thorne, J.; and Vlachos, A. 2018.
\newblock Automated Fact Checking: Task Formulations, Methods and Future
  Directions.
\newblock In \emph{Proceedings of the 27th International Conference on
  Computational Linguistics}, 3346--3359. Santa Fe, New Mexico, USA: ACL.

\bibitem[{Wang et~al.(2021{\natexlab{a}})Wang, Tang, Duan, Wei, Huang, Ji, Cao,
  Jiang, and Zhou}]{wang-etal-2021-k}
Wang, R.; Tang, D.; Duan, N.; Wei, Z.; Huang, X.; Ji, J.; Cao, G.; Jiang, D.;
  and Zhou, M. 2021{\natexlab{a}}.
\newblock {K-Adapter}: {I}nfusing {K}nowledge into {P}re-{T}rained {M}odels
  with {A}dapters.
\newblock In \emph{Findings of the Association for Computational Linguistics:
  ACL-IJCNLP 2021}, 1405--1418. Online: ACL.

\bibitem[{Wang(2017)}]{wang-2017-liar}
Wang, W.~Y. 2017.
\newblock {``}Liar, Liar Pants on Fire{''}: A New Benchmark Dataset for Fake
  News Detection.
\newblock In \emph{Proceedings of the 55th Annual Meeting of the ACL (Volume 2:
  Short Papers)}, 422--426. Vancouver, Canada: ACL.

\bibitem[{Wang et~al.(2021{\natexlab{b}})Wang, Gao, Zhu, Zhang, Liu, Li, and
  Tang}]{kepler}
Wang, X.; Gao, T.; Zhu, Z.; Zhang, Z.; Liu, Z.; Li, J.; and Tang, J.
  2021{\natexlab{b}}.
\newblock KEPLER: A Unified Model for Knowledge Embedding and Pre-trained
  Language Representation.
\newblock \emph{Trans. Assoc. Comput. Linguistics}, 9: 176--194.

\bibitem[{Zellers et~al.(2019)Zellers, Holtzman, Rashkin, Bisk, Farhadi,
  Roesner, and Choi}]{NEURIPS2019_3e9f0fc9}
Zellers, R.; Holtzman, A.; Rashkin, H.; Bisk, Y.; Farhadi, A.; Roesner, F.; and
  Choi, Y. 2019.
\newblock Defending Against Neural Fake News.
\newblock In \emph{Advances in Neural Information Processing Systems},
  volume~32. Curran Associates, Inc.

\bibitem[{Zhang et~al.(2019)Zhang, Han, Liu, Jiang, Sun, and
  Liu}]{zhang-etal-2019-ernie}
Zhang, Z.; Han, X.; Liu, Z.; Jiang, X.; Sun, M.; and Liu, Q. 2019.
\newblock {ERNIE}: Enhanced Language Representation with Informative Entities.
\newblock In \emph{Proceedings of the 57th Annual Meeting of the Association
  for Computational Linguistics}, 1441--1451. Florence, Italy: ACL.

\end{thebibliography}

\end{document}